\documentclass[twocolumn]{article}

\usepackage{arxiv}

\usepackage[utf8]{inputenc} 
\usepackage[T1]{fontenc}    
\usepackage{hyperref}       
\usepackage{url}            
\usepackage{booktabs}       
\usepackage{amsfonts}       
\usepackage{nicefrac}       
\usepackage{microtype}      
\usepackage{lipsum}
\usepackage{graphicx}

\usepackage{amssymb}
\usepackage{latexsym}

\usepackage{url}
\usepackage{xcolor}
\usepackage{svg}

\usepackage{amsmath}
\usepackage{float}
\usepackage{wrapfig}
\usepackage{fancyhdr}
\usepackage{subcaption}
\usepackage{bm}
\usepackage{tabularx}
\usepackage{float}

\usepackage{authblk}

\usepackage{setspace}          

\usepackage{microtype}

\usepackage{hyperref}

\hypersetup{
    colorlinks=true,
    linkcolor=blue,
    filecolor=magenta,      
    urlcolor=blue,
    pdftitle={Overleaf Example},
    pdfpagemode=FullScreen,
    }

\usepackage[capitalize]{cleveref}
\usepackage{natbib}

\definecolor{newcolor}{rgb}{.8,.349,.1}

\title{InstanSeg: an embedding-based instance segmentation algorithm optimized for accurate, efficient and portable cell segmentation}

\usepackage{authblk} 

\author[1]{Thibaut Goldsborough}
\author[1]{Ben Philps}
\author[2]{Alan O'Callaghan}
\author[2]{Fiona Inglis}
\author[2]{Leo Leplat}
\author[3]{Andrew Filby}
\author[1]{Hakan Bilen}
\author[2,4]{Peter Bankhead}

\affil[1]{School of Informatics, University of Edinburgh, UK}
\affil[2]{Centre for Genomic and Experimental Medicine, Institute of Genetics and Cancer, University of Edinburgh, UK}
\affil[3]{Flow Cytometry Core Facility, Newcastle University, UK}
\affil[4]{Edinburgh Pathology and CRUK Scotland Centre, Institute of Genetics and Cancer, The University of Edinburgh, UK}

\date{} 

\usepackage{cuted}
\usepackage{lipsum}
\begin{document}

\begin{strip}
  \vspace*{\dimexpr-\baselineskip-\stripsep\relax}
  \centering
  \maketitle

  \vskip\baselineskip
\noindent\makebox[\textwidth]{\rule{1.1\paperwidth}{0.4pt}}
  \vskip\baselineskip
\end{strip}



\textbf{\textit{Abstract} - Cell and nucleus segmentation are fundamental tasks for quantitative bioimage analysis. Despite progress in recent years, biologists and other domain experts still require novel algorithms to handle increasingly large and complex real-world datasets. These algorithms must not only achieve state-of-the-art accuracy, but also be optimized for efficiency, portability and user-friendliness. Here, we introduce InstanSeg: a novel embedding-based instance segmentation pipeline designed to identify cells and nuclei in microscopy images. Using six public cell segmentation datasets, we demonstrate that InstanSeg can significantly improve accuracy when compared to the most widely used alternative methods, while reducing the processing time by at least 60\%. Furthermore, InstanSeg is designed to be fully serializable as TorchScript and supports GPU acceleration on a range of hardware. We provide an open-source implementation of InstanSeg in Python, in addition to a user-friendly, interactive QuPath extension for inference written in Java. Our code and pre-trained models are available at this \href{https://github.com/instanseg/instanseg}{https URL}.}
\section{Introduction}

Quantitative cell biology and digital pathology frequently involve analysing cell properties from imaging data. This includes the size, shape, location and staining of cells, as well as tissue-level properties such as cell counts and interactions. Determining features from individual cells relies on accurate detection and boundary identification. Detecting stained nuclei is typically used as a proxy for segmenting entire cells, especially in brightfield or fluorescence images where cell boundaries may not be visible. Despite its ubiquity, the task of identifying nucleus boundaries from microscopy images has proven deceptively difficult \citep{meijering_cell_2012}, and remains a very active area of research, including though grand challenges and contests \citep{caicedo_nucleus_2019}. The difficulty of the task derives from the immense variation seen in biological images, combined with the specific problem of distinguishing nuclei that may be densely packed and therefore have indistinct boundaries. The overall challenge fits in the more general computer vision field of instance segmentation. 

The goal of instance segmentation is to assign pixels to individual instances. Most current methods can be categorized as either proposal-based and proposal-free methods. Proposal-based methods, such as Mask-R-CNN \citep{he_mask_2018}, Fast-R-CNN \citep{girshick_fast_2015} or more recently SAM \citep{kirillov_segment_2023}, typically rely on a Deep Neural Network (DNN) to predict bounding boxes for every instance, followed by a second network that predicts a dense binary mask for every bounding box. These two-step approaches typically come at a high computational cost and slower inference speeds, and are seldom used in end-to-end implementations by biologists. An exception to this is the popular Stardist \citep{schmidt_cell_2018} method, which relies on the prediction of a more detailed, usually 32-sided bounding polygon. This circumvents the need for the second-step binary mask prediction, at the cost of some fine-grained detail at the object boundary. 

The alternative to these approaches are the proposal-free methods, which usually rely on a Fully Convolutional Network (FCN) predicting dense feature maps that are later post-processed to resolve individual instances. Segregating foreground from background pixels is relatively standard across methods and relies on the prediction of a foreground probability map that can be thresholded. The main differences between current methods relate to how they separate touching or overlapping instances. One popular approach relies on the prediction of an instance-boundary distance map \citep{naylor_segmentation_2019,greenwald_whole-cell_2022}. Other methods use two dimensional offsets \citep{graham_hover-net_2019} or flows \citep{stringer_cellpose_2021} to the instance centroid. These serves as input to a watershed or flow tracking algorithm, which is usually seeded at local maxima in the foreground probability map. In practice, it is difficult for any predictor to place exactly one seed for every instance, which can is crucial to avoid errors in the segmented output. Furthermore, the computational cost of a watershed transform commonly causes a performance bottleneck. Especially in the field of computational pathology, where a single whole slide image (WSI) is typically 10-40 GB in size and may depict over a million cell instances, computational efficiency is a crucial practical consideration. 

More recent work by \cite{neven_instance_2019} and its adaptation to 2D and 3D microscopy images as \textit{EmbedSeg} \citep{lalit_embedseg_2022}, introduces a novel and powerful approach to proposal-free, boundary-detailed instance segmentation. The method builds on an Erfnet \citep{romera_erfnet_2018} backbone to generate dense pixel embeddings that are further used to cluster pixels of the same instance while segregating pixels from neighbouring instances. Despite the potential of embedding-based segmentation methods, EmbedSeg has received relatively little attention in the field. This is likely due to (1) a dependence on the Erfnet backbone, (2) non-commercial restrictions on code reuse, (3) postprocessing steps that cannot easily be ported outside of the python environment, (4) a seed sampling strategy that differs between train and test time, (5) a restrictive assumption on the distribution of embeddings around sampled seeds. Together, these hinder the accuracy, efficiency and portability of the method, and have prevented its integration both in commercial applications and in user-friendly open-source software packages widely used by biologists, such as Fiji \citep{schindelin_fiji_2012} or QuPath \citep{bankhead_qupath_2017}.

To address these issues, we present \textit{InstanSeg}, an novel embedding-based instance segmentation method optimized for accuracy, efficiency, and portability. Building on a modified U-Net backbone \citep{ronneberger_u-net_2015}, InstanSeg uses a lightweight neural network to cluster pixel embeddings around optimally selected seeds. Our novel approach sets a new state of the art in terms of accuracy on six public nucleus segmentation datasets, while reducing the processing time by a factor of approximately 2.5 -- 45x compared to the most widely used current methods. Furthermore, InstanSeg is a highly vectorized pipeline that can be efficiently used for inference on a laptop GPU and serialized in a self-contained TorchScript implementation. This strategy means that InstanSeg can be used not only in Python, but also integrated into software tools written in other languages, without a requirement to replicate any complex pre- and post-processing steps.

\section{InstanSeg, a novel embedding-based segmentation algorithm}\label{subsec4_2}

\begin{figure*}
    \centering
    \includegraphics[width=0.6\linewidth]{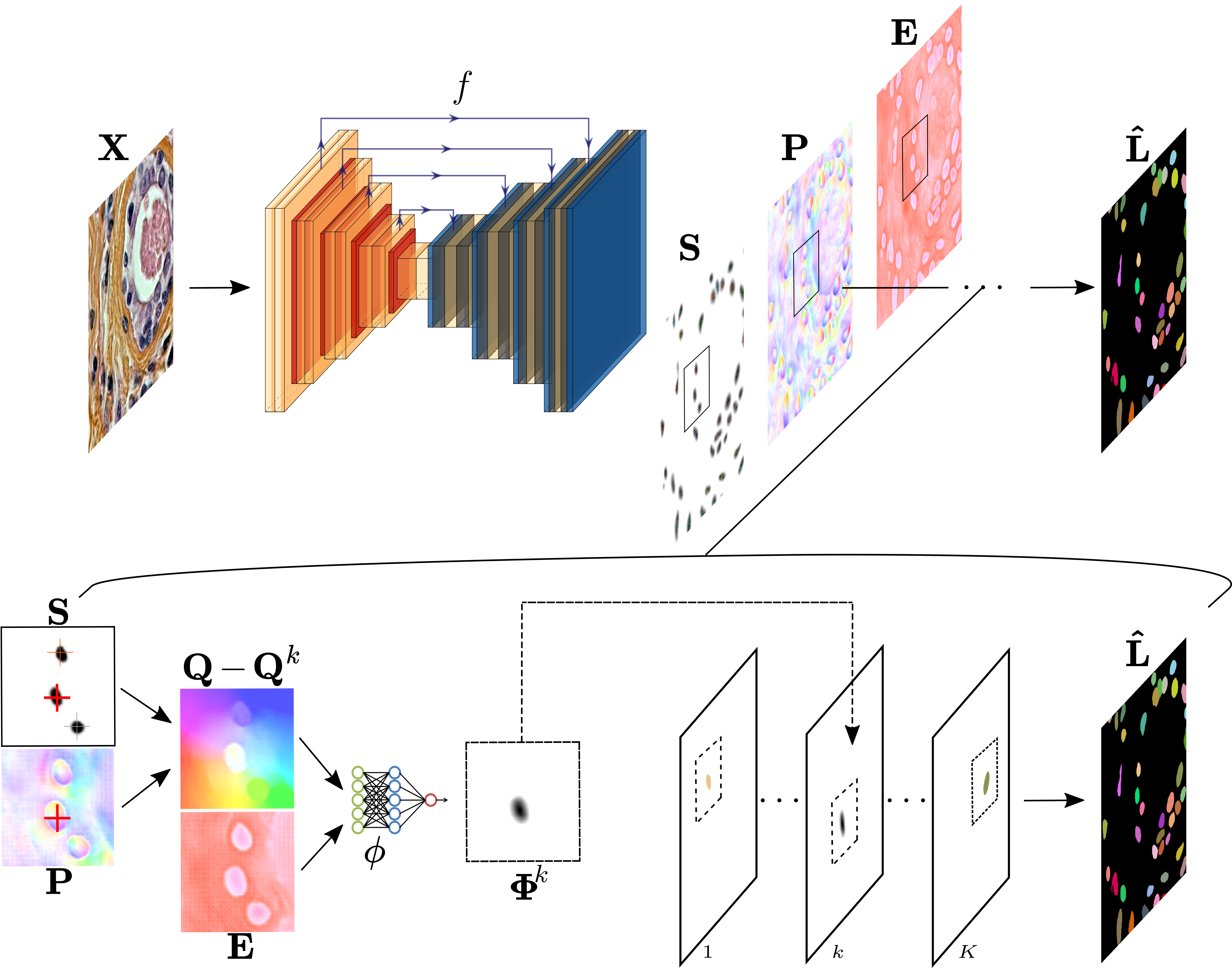}
    \caption{The InstanSeg pipeline. A feature encoder $f$ and three feature heads $s,p$ and $e$ transform an input image $\bm{X}$ into a seed map $\bm{S}$, a positional embedding $\bm{P}$ and a conditional embedding $\bm{E}$. We sample local maxima in the seed map to find the coordinates of seed pixels $\bm{u}^k$ (e.g red cross) and compute the relative offsets between the positional embeddings and each seed embedding $\bm{Q}-\bm{Q}^k$. These offsets, along with conditional embeddings $\bm{e}$ serve as input to the instance segmentation head $\Phi$, which outputs a probability map $\bm{\Phi}^k$ of each pixel belonging to instance $k$. The final labelled segmentation map $\bm{\hat{L}}$ is obtained by merging the K probability maps. Input image is from the TNBC 2018 dataset \citep{naylor_segmentation_2019}.}
    \label{fig:instanseg-main}
\end{figure*}

\subsection{Problem setting}

Our goal is to learn an instance segmentation model that takes an input image $\bm{X}$ of shape $C \times H \times W$ and predicts a labelled segmentation map $\bm{\hat{L}}$ of shape $H \times W$ which not only distinguishes instances from background but also individual neighbouring instances from each other. $C$ denotes the number of image channels. Given a dataset $\mathcal{D}$ with $|D|$ doublets, each including an image $\bm{X} \in \mathbb{R}^{C\times H\times W}$ and an instance segmentation mask $\bm{L} \in \mathbb{R}^{H\times W}$, we would like to train the parameters of our model to estimate the ground truth $\bm{L}$. Each element of $\bm{L}$ is a discrete value between $1$ and $K$ indicating that the corresponding pixel belongs to one of $K$ instances in the image.

\subsection{Intuition}

We treat the task as a pixel assignment problem, where we want to associate each pixel $\mathbf{X}_{ij}$ at the coordinates $(i,j)$ to its corresponding object. As a proxy, we seek a set of \textit{seed} pixels (e.g. centroids) to represent each object, and then relate each pixel to these seeds. A simple approach is to use a similarity metric to relate pixels to seeds. However, this approach runs into two problems: (1) fully convolutional networks are translationally invariant, and may struggle to segregate pixels that are situated far apart on the image plane. (2) A similarity metric does not obviously relate to a probability of belonging to an object, especially when considering objects that are overlapping or of vastly different sizes. 

The first problem can be solved by introducing a pixel coordinate system either inside the model \citep{kulikov_instance_2020} or to the model outputs \citep{neven_instance_2019}; we favor the latter for ease of implementation. For the second problem, we need to find a function that maps a similarity metric to a probability of belonging to an object. Ideally, the function should be conditioned on higher order object properties, such as size, orientation or shape. Previous work by \cite{neven_instance_2019} uses a Gaussian function to map a similarity metric to a probability of belonging to an instance. While intuitive, this formulation enforces pixel embeddings to lie in a circular or elliptical distribution around the instance seed, a task that can be difficult when segmenting touching non-convex objects that do not have obvious centres. Instead, we do not seek a specific functional form, and opt to instantiate it as a neural network instead, thereby relaxing the underlying assumptions on the distribution of embeddings around instance seeds. 

Hence, we solve the task in three steps, we (1) seek suitable seed pixels to represent each object, (2) compute a similarity metric to relate each pixel to each seed pixel, (3) use a neural network to map a similarity metric, conditioned on learnt higher object properties, to a probability of belonging to an object. 


\subsection{Our Method}

As illustrated in \cref{fig:instanseg-main}, we build our model with a feature encoder $f$ and three auxiliary prediction heads: $s$, $p$ and $e$ predicting the location of seed pixels $\bm{S}$, positional embeddings $\bm{P}$ and conditional embeddings $\bm{E}$ respectively. A fourth prediction head $\phi$ is finally used to predict instance labels.
The feature encoder $f$ takes in a $C$-dimensional $H\times W$ input image and outputs a feature map $\bm{F}$ consisting of $D_f$ feature maps, each with $H\times W$ spatial resolution.
The auxiliary encoders 
$p$
and 
$e$
take in the feature map $\bm{F}$ and output $D_p$ and $D_e$ feature maps with $H\times W$ spatial dimensions respectively. 

\begin{figure}[!t]
    \centering
    \includegraphics[width=1\linewidth]{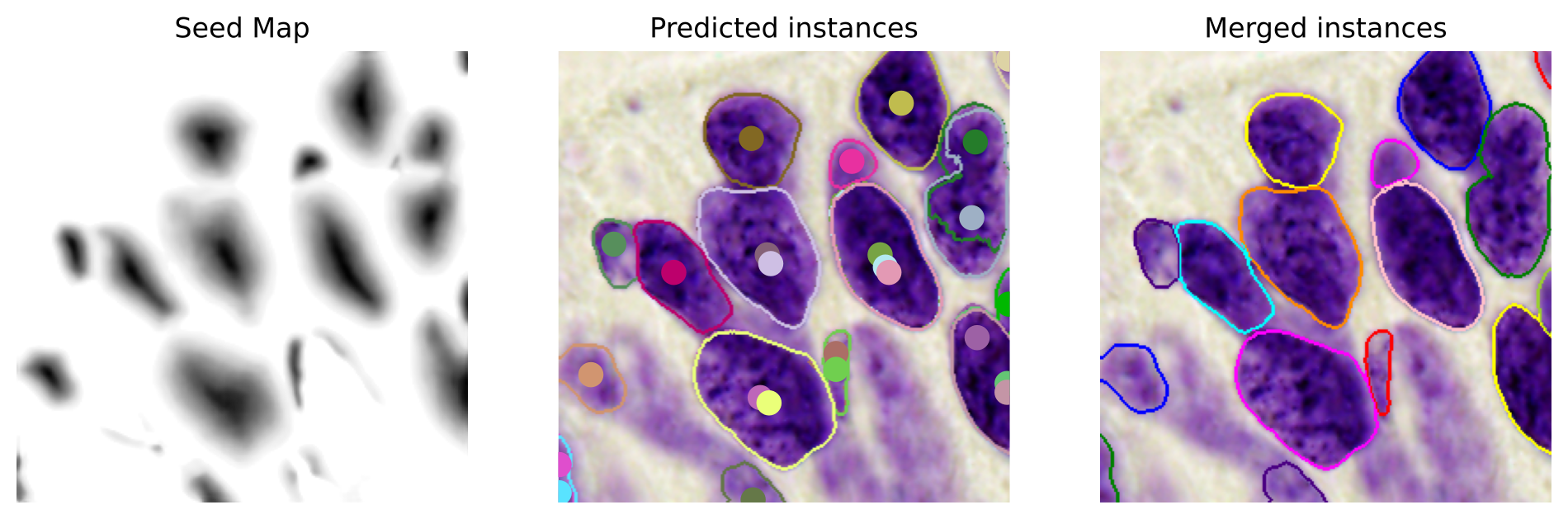}
    \caption{InstanSeg uses a local maxima algorithm on a predicted seed map (left image) to locate suitable seed(s) for each object. In large instances, there may be multiple seeds detected within the object. InstanSeg was trained to predict the full instance irrespective of the sampled seed location, as a result predicted objects corresponding to the same instance greatly overlap (centre image). We merge objects with large overlaps by taking their union (right image). Input image from the DSB 2018 dataset \citep{caicedo_nucleus_2019}.}
    \label{fig:postprocessing}
\end{figure}

\subsubsection{Seed selection}

The aim of the seed head is to predict the location of suitable seed pixels that best represent an instance. 
Previous embedding-based segmentation methods have used the centroid \citep{neven_instance_2019} or medoid \citep{lalit_embedseg_2022} of the object's positional embeddings as their seed location. While sensible, these are not obvious to sample at test time when objects are not yet resolved -- requiring a sampling strategy that differs between the training and testing phases, and also necessitates an additional term in the loss computation.

We seek a simpler more consistent seed sampling strategy. We hypothesise that pixels that lie close to the instance centre are suitable seed pixels and hence train the seed head to predict for each foreground pixel, the distance to the instance boundary. Hence, we minimize the following loss function:
\begin{equation}
    \label{eq:lossseg}
    \mathcal{L}_{s} = \frac{1}{|D|} \sum_{(\bm{X},\bm{L})\in \mathcal{D}} \sum_{i=1}^{H} \sum_{j=1}^{W} \ell_{1}(\bm{S}_{ij},d(\bm{L}_{ij}))
\end{equation} where the subscript $i,j$ denotes the feature at the spatial location $(i,j)$, $d(\bm{L}_{ij})$ denotes the relative distance from the pixel location $(i,j)$ to the nearest instance boundary, and $\ell_{1}$ is the $L1$ loss function.

Importantly, we use the same seed sampling strategy during both training and testing: we find the location $\bm{u}^k$ of the seed pixel belonging to instance $k$ by sampling local maxima in the seed map $\bm{S}$. Note that there may be more than one seed pixel per instance, which can easily be resolved in postprocessing. This approach allows for larger instances to be segmented by merging smaller overlapping fragments (see top-right nucleus in \cref{fig:postprocessing} for an example).

\subsubsection{Similarity metric computation}

We use the computed coordinate for each seed to obtain a seed embedding by using the auxiliary positional encoder $p$, \textit{i.e.} $\bm{P}^k= \bm{P}_{\bm{u}^k} $ where $\bm{P}_{\bm{u}^k}$ denotes $D_p$ dimensional encoding at the location $\bm{u}^k$. We then compare each embedding pixel $\bm{P}_{ij}$ to each sampled seed embedding $\bm{P}^k$ to relate pixels to their corresponding objects. To this end, we calculate the offset between $\bm{P}_{ij}$ and $\bm{P}^k$. Specifically, we compute $\bm{Q}_{ij}-\bm{Q}^k=(\bm{P}_{ij} + \bm{O}_{ij})-(\bm{P}^k + \bm{O}_{\bm{u}^k})$ where $\bm{O}$ denotes linear coordinates in the H and W dimensions. Unlike previous embedding-based methods, we do not limit the dimensionality of the coordinate system $\bm{O}$ to match the input image dimensionality -- specifically, when $D_e > 2$, we match the dimension of $\bm{O}$ by adding empty channels, allowing for embeddings $\bm{Q}_{ij}$ to lie outside of the image plane. We hypothesize that the increased embedding space enables better separation of crowded instances.

\subsubsection{Instance probability mapping using a neural network}

For each seed embedding $\bm{P}^k$, we use a separate head $\phi$ to predict a binary probability map $\bm{\Phi}^k$ of each pixel belonging to the instance in which $\bm{u}^k$ resides. A simple segmentation head could hence take pixel offsets $\bm{Q}_{ij}-\bm{Q}^k$ as sole input. However, learnt pixel offsets may not be sufficient to separate overlapping or crowded instances. For example, consider a small instance bordering a much larger one, pixel offsets near their boundary will tend to be closer to the smaller instance's seed and lead to a biased pixel assignment. To address this, we use pixel embeddings $\bm{E}_{ij}$ from a separate encoder $e$ as an additional input to the segmentation head, and hypothesize that these can encode higher order object properties such as orientation, shape or size.

The instance segmentation head $\phi:\mathbb R^{D_e+ D_p} \to \mathbb R^1$ takes in $\bm{Q}_{ij}-\bm{Q}^k$ along with $\bm{E}_{ij}$ and outputs a scalar $\bm{\Phi}_{ij}^k$. The head $\phi$ is instantiated as a multi layer perceptron (MLP) with a single hidden layer and is trained by minimizing the loss

\begin{equation}
    \label{eq:lossins}
    \mathcal{L}_{i} = \frac{1}{|D|} \sum_{(\bm{X},\bm{L})\in \mathcal{D}} \sum_{k=1}^{K}\sum_{i=1}^{H} \sum_{j=1}^{W} \ell_{lh}(\bm{\Phi}_{ij}^k,\bm{L}_{ij}^k),
\end{equation} where $\ell_{lh}$ is the Lovasz-hinge loss \citep{berman_lovasz-softmax_2018}, and the scalar $\bm{L}_{ij}^k$ is given by 

\[
\bm{L}_{ij}^k = 
\begin{cases} 
1 & \text{if } \bm{L}_{ij} = k \\
0 & \text{if } \bm{L}_{ij} \neq k .
\end{cases}
\]
Finally by combining \cref{eq:lossseg} and \cref{eq:lossins}, the whole segmentation model is trained by minimizing the following term jointly:

$$
\mathcal{L} = \mathcal{L}_{s} + \mathcal{L}_{i}.
$$

In the interest of computational efficiency, we only consider pixels in the vicinity of the sampled seed locations, as illustrated in the pipeline diagram. As a result, the memory requirement of postprocessing the model outputs $\bm{F}$ to a labelled segmentation map $\bm{\hat{L}}$ scales linearly with the number of instances, independently of image size.

\subsubsection{Inference}
We use an identical seed sampling strategy as in our training phase. There may be multiple local maxima sampled within a single object; this is difficult to avoid, as in general it is hard for any predictor to place exactly one seed inside each object. 
However, the network $\phi$ was trained to output the same probability map $\bm{\Phi}^k$ for any seed embedding $\bm{P}^k$ that was sampled within the instance $k$. As a result, we can identify redundant probability maps $\bm{\Phi}^k$ using an intersection over union metric. We merge redundant $\bm{\Phi}^k$ by taking their union, see \cref{fig:postprocessing} for an illustration. Finally, we obtain a labelled segmentation map using  \

\begin{equation}
    \mathbf{\hat{L}}_{ij} = 
    \begin{cases} 
    {\arg\max}_{(k)} \bm{\Phi}_{ij}^k & \text{if } {\max}_{(k)} \bm{\Phi}_{ij}^k \geq 0  \\
    0 & \text{otherwise}.
    \end{cases}
\end{equation}

\begin{table*}[!h]
\centering
\caption{ \label{tab:datasets} Total number of images and instances for each dataset.}
\footnotesize
\begin{tabular}{|c|c|c|}
\hline
Dataset& Images (Train/Val/Test) & Instances (Train / Val / Test ) \\
\hline
CoNSeP \citep{graham_hover-net_2019}     & 21 / 6 / 14   & 12640 / 2915 / 8777\\
TNBC 2018 \citep{jack_extention_2021}  & 54 / 7 / 7    & 4434 / 398 / 349 \\
MoNuSeg  \citep{kumar_dataset_2017}   & 29 / 8 / 14   & 18195 / 5978 / 6699\\
LyNSeC   \citep{hussein_lynsec_2023}    & 559 / 70 / 70 & 121436 / 13910 / 16433\\
NuInsSeg  \citep{mahbod_nuinsseg_2024}  & 532 / 66 / 67 & 24570 / 3219 / 2909 \\
IHC TMA  \citep{wang_simultaneously_2024}    & 212 / 27 / 27 & 7284 / 971 / 906\\
\hline
\end{tabular}
\end{table*}

\subsection{Backbone network architecture} \label{sec:backbone}

We reviewed the accuracy and performance of a number of publicly available convolutional and transformer based model architectures (results not shown). We found the best performing architecture was the UNet \cite{ronneberger_u-net_2015} based implementation from Cellpose \citep{stringer_cellpose_2021}, which uses maxpooling in the encoder blocks and nearest neighbour interpolation in the decoder blocks, with residual connections in each block. The decoder layers use summation instead of concatenation to merge skip connections.  We therefore adopt a modified version of Cellpose's backbone network. Specifically, we remove the style vectors to improve translational invariance and we change the ordering of operations inside each block to convolution - normalisation - activation, as default in other libraries. Our resulting backbone model has approximately four million parameters, around half the number of trainable parameters compared to the original Cellpose model, with similar accuracy.

\subsection{Training details}

InstanSeg is implemented using the Pytorch library. We train over 500 epochs, each consisting of 1000 batches of size 3. Each image is a random $256 \times 256$ pixel crop. We use a fixed learning rate of $0.001$ and use the Adam optimizer \citep{kingma_adam_2017}, we store model weights when the highest scoring $F_1^\mu$ on the validation set is reached. 

We use an additional pretraining step of 10 epochs where we substitute L1 regression of the distance map with binary cross entropy in \cref{eq:lossseg} and replace the Lovasz-hinge loss with Dice loss in \cref{eq:lossins}. We find that this greatly accelerates and stabilizes convergence in the initial stages of training. During training, we cap the number of instances in \cref{eq:lossins} to $K = 50$ for lower GPU memory requirements and faster convergence. 

To fairly compare InstanSeg with other methods, we do not resize the input images for the benchmark results. However, in the models we make publicly available, we use the known image resolution in terms of µm/px to standardize the scale the training images. This solves the problem of how to resize images acquired at different magnifications to be compatible with the trained model.

Our code and models are available at this \href{https://github.com/instanseg/instanseg}{https URL} \footnote{https://github.com/instanseg/instanseg}.

\subsection{Augmentations}

For the benchmarks, we use minimal augmentations during training so as to fairly compare results with other methods.  These include horizontal and vertical flips, axis-aligned rotations and random crops of size $256 \times 256$ pixels. For additional models we make publicly available, we use additional augmentations including stain normalization, hue, brightness and contrast shifts.

\subsection{Tiled Predictions}

The GPU memory requirements of InstanSeg are low enough for each of the validation and test set images without having to tile or resize the input images, therefore we only pad the input images so that the height and width are divisible by 32 as required by our U-Net backbone. For inference on larger images, such as whole slide images, we run InstanSeg on individual tiles to obtain labeled images with a fixed overlap of 80 pixels and merge the predicted labels by matching duplicated objects using an IoU metric. This differs from other tiling approaches that merge the intermediate model outputs and postprocess the resulting stitched image. Our approach enables InstanSeg to run on WSIs that are bigger than system RAM and can benefit from GPU acceleration for postprocessing model outputs.

\subsection{Test Time Augmentations}

Test time augmentations (TTA) involve running a model on a set of augmented versions of a test image and aggregating the model predictions to increase model accuracy at the expense of compute time and/or memory. TTA is an expensive operation that is seldom used in real world applications, but is commonly reported in segmentation benchmarks. To fairly compare with other methods, we report both results with and without TTA in our benchmarks. Our TTA implementation uses ttach \citep{iakubovskii_qubvelttach_2024}, and involves 16 combinations of axis aligned rotations and flips. We cannot directly pool the model outputs, as these are orientation dependent. Instead, we merge the outputs of our segmentation head $\bm{\Phi}^k$ using element wise median pooling. Unless explicitly specified, we do not use TTA when reporting our results.

\begin{table*}[h!]
\caption{\label{tab:maintable} Quantitative segmentation results on 6 publicly available datasets. Best results are shown in bold, second best results in italics.}
\centering
\footnotesize
\begin{tabular}{|c|cc|cc|cc|cc|cc|cc|}
\hline
& \multicolumn{2}{c|}{TNBC 2018}   & \multicolumn{2}{c|}{NuInsSeg}   & \multicolumn{2}{c|}{MoNuSeg}   & \multicolumn{2}{c|}{IHC TMA}   & \multicolumn{2}{c|}{CoNSeP}   & \multicolumn{2}{c|}{LyNSeC}   \\
\hline
                 & $F_1^{\mu}$    & $F_1^{0.5}$    & $F_1^{\mu}$    & $F_1^{0.5}$    & $F_1^{\mu}$    & $F_1^{0.5}$    & $F_1^{\mu}$                      & $F_1^{0.5}$                     & $F_1^{\mu}$                    & $F_1^{0.5}$                    & $F_1^{\mu}$                   & $F_1^{0.5}$                   \\
\hline
 StarDist        & 0.645          & 0.896          & 0.494          & \textit{0.799} & 0.543          & 0.846          & 0.470                            & 0.798                           & 0.418                          & 0.690                          & 0.701                         & 0.920                         \\
 HoVer-Net       & 0.546          & 0.768          & 0.374          & 0.635          & 0.438          & 0.707          & 0.304                            & 0.559                           & 0.312                          & 0.538                          & 0.659                         & 0.886                         \\
 CellPose        & 0.627          & 0.835          & 0.497          & 0.788          & 0.553          & 0.850          & 0.545                            & 0.811                           & 0.389                          & 0.626                          & 0.701                         & 0.911                         \\
 EmbedSeg (TTA)  & 0.641          & 0.870          & 0.492          & 0.761          & 0.560          & 0.853          & -                           & -                           & 0.434                          & 0.619                          & -                         & -                         \\
\hline
 InstanSeg       & \textit{0.699} & \textit{0.898} & \textit{0.511} & 0.787          & \textbf{0.575} & \textbf{0.861} & \textit{0.560}                   & \textbf{0.836}                  & \textit{0.483}                 & \textit{0.706}                 & \textit{0.724}                & \textbf{0.923}                \\
 InstanSeg (TTA) & \textbf{0.704} & \textbf{0.899} & \textbf{0.527} & \textbf{0.805} & \textit{0.566} & \textit{0.860} & \textbf{0.573}                   & \textit{0.834}                  & \textbf{0.491}                 & \textbf{0.709}                 & \textbf{0.725}                & \textit{0.922}                \\
\hline
\end{tabular}
\end{table*}

\section{Baselines, Experiments and Results}

\subsection{Datasets}

We benchmark InstanSeg on six independent publicly available datasets. We focus on datasets with clearly-defined licensing terms to facilitate reuse \textit{TNBC 2018} \citep{jack_extention_2021}, \textit{NuInsSeg} \citep{mahbod_nuinsseg_2024}, \textit{IHC TMA} \citep{wang_simultaneously_2024}, \textit{CoNSeP} \citep{graham_hover-net_2019}, \textit{MoNuSeg} \citep{kumar_dataset_2017} and \textit{LyNSeC} \citep{hussein_lynsec_2023}. We report summary dataset statistics in \cref{tab:datasets}.

We do not include the DSB 2018 dataset \citep{caicedo_nucleus_2019} for three main considerations, (1) the architecture and rationale behind building InstanSeg was extensively developed using this dataset and reporting results unfairly benefits our method over the other baselines, (2) an abnormally high number of labelling mistakes and (3) the lack of pixel resolution information, limiting the use of this dataset in real world applications.

\subsection{Evaluation Metrics}

The $F_1$ score was used as a metric for detection accuracy. Predicted objects having an IoU with a ground-truth object greater than the IoU threshold $\tau$ is considered a true positive $T_p$, while if the IoU is smaller than $\tau$, it is considered a false positive $F_P$. The number of false negatives $F_N$ is calculated as the difference between the number of ground truth objects and the number of true positives. The score is calculated as $F_1 =\frac{2T_P}{2T_P+F_P+F_N}$. We report both $F_1^{0.5}$, determined at $\tau=0.5$ and $F_1^\mu$, calculated as the mean $F_1$ score over the interval $[0.5,0.9]$ with a step of $0.1$. Our metrics are calculated using the Stardist implementation \footnote{https://github.com/stardist/stardist}.

\subsection{Baselines}

For evaluating \textit{InstanSeg}, we selected the three most widely-used nucleus and cell detection methods for bioimage analysis based on deep learning, along with a fourth method that is the most similar to our proposed approach. These methods are widely reported to represent the current state of the art, and cover a range of different segmentation approaches. 

We do not report previously published results because exact datasets, training splits and evaluation metrics differ among the literature. We retrain all methods from scratch on identical train, validation and test splits on all six datasets. For all baselines, we use the official code and default hyper-parameters when possible. Computational restraints prevent us from optimizing training hyperparameters (e.g. batch size, learning rate, regularization) on the individual datasets, as some of the methods required multiple days of training. As a result we use default training parameters for all methods and datasets. We acknowledge that this might not fully exploit the potential of each method and lead to some of the methods under-performing. To improve fairness, we optimize InstanSeg's hyperparameters on a separate dataset (DSB 2018) and refrain from including this dataset in our results.

\textbf{Cellpose} \citep{stringer_cellpose_2021} is a popular whole cell and nucleus segmentation pipeline. The method is based on a modified UNet backbone trained to predict a simulated heat diffusion pattern initiated at instance centroids. Despite being computationally expensive, the method is known to achieve high accuracies for both nucleus and whole cell segmentation, and was originally intended as a generalist algorithm that performs well over a wide range of image modalities and image resolutions. We run Cellpose with default parameters, on all three (RGB) input channels.

\textbf{Stardist} \citep{schmidt_cell_2018}  uses a modified UNet backbone to predict a vector for every pixel in an image. During training, the vector is regressed to the distance to the object boundary along a set of predefined angles termed radial directions. Following a non-maximal suppression step, the method predicts a star-convex polygon approximating every instance. We use N = 32 radial dimensions.

\textbf{HoVer-Net} \citep{graham_hover-net_2019}, is an encoder-decoder network predicting horizontal and vertical distances to the instance centre of mass which are then post-processed using a marker controlled watershed on the predicted gradients. The method was originally intended to both segment and classify nuclei, but we only consider the segmentation branch of the method. For timing Hover-Net, we use the Monai implementation and set the postprocessing mode to ``Fast". We use the default sliding window inference function using a batch size of 8.

\textbf{EmbedSeg} \citep{lalit_embedseg_2022} is a recent adaption of \citep{neven_instance_2019} to microscopy images and is conceptually the method most similar to the proposed InstanSeg. We set n\_sigma to 5 and keep all parameters as default. Note that were unable to produce meaningful segmentation results on the IHC TMA and LyNSeC datasets without modification of the source code. It was unclear why the method failed on these datasets, but manually changing the values of $n_x$ and $n_y$ parameters at test time improved results. We omit EmbedSeg results on these two datasets. We use test-time augmentations (TTA) when reporting the accuracy results, but disable TTA when reporting time efficiency of the method. When reporting time efficiency, we set the postprocessing mode to ``Fast". 

\begin{figure*}
    \centering
    \includegraphics[width=1\linewidth]{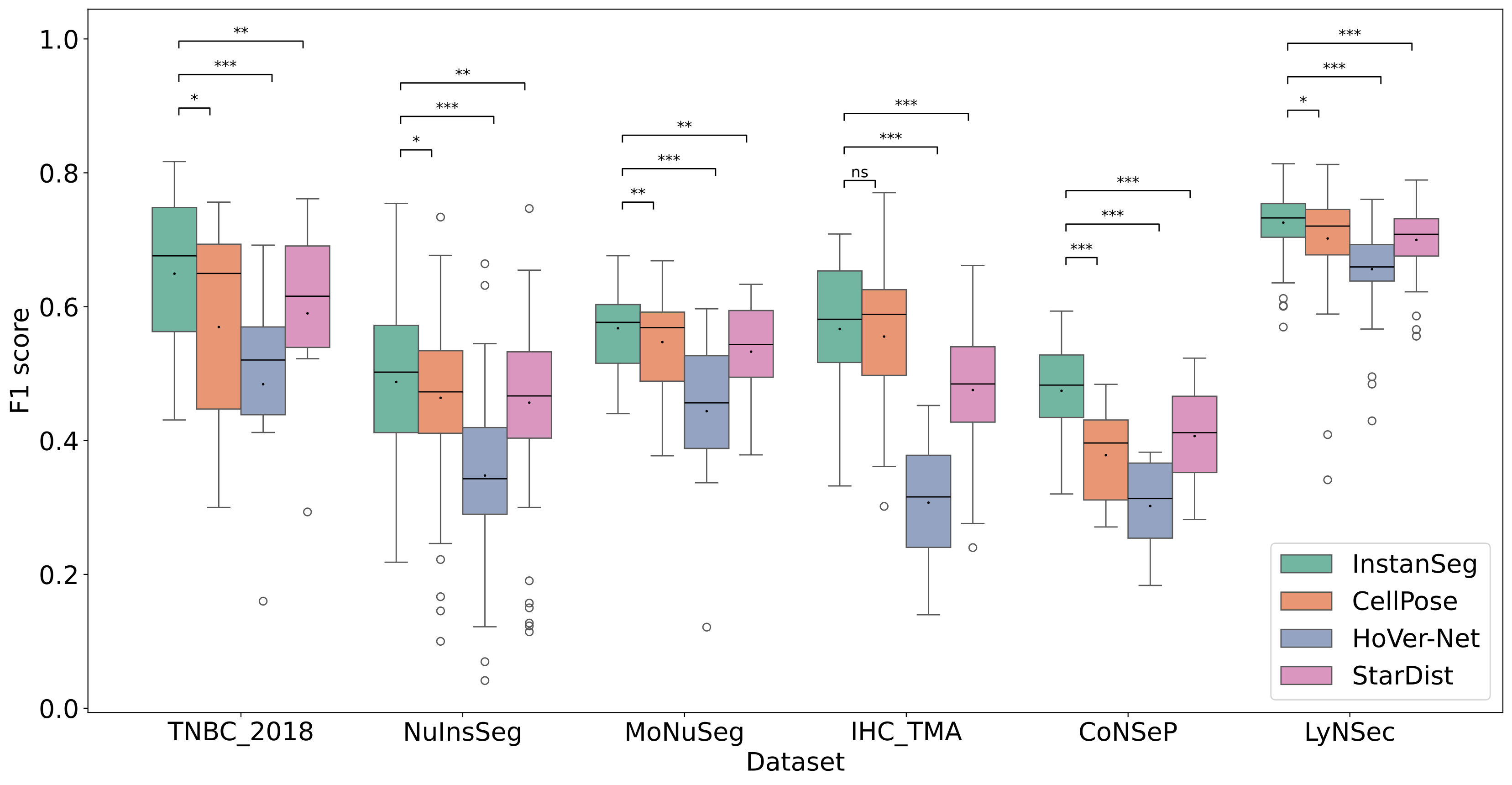}
    \caption{Box plot showing the distribution of $F_1^{\mu}$ scores on the entire testing set for each of the six datasets. We compare InstanSeg against each of the three other methods using a paired t-test, and report the significance level of the p-value.}
    \label{fig:boxplot}
\end{figure*}

\section*{Results and discussion}

We benchmark InstanSeg on six independent public nucleus segmentation datasets and report our results in \cref{tab:maintable} , we further display per-image F1 scores in \cref{fig:boxplot}. We find that InstanSeg's performance is consistently superior to the previous state of the art nucleus segmentation pipelines. Our base InstanSeg method surpasses all previous methods on 11/12 metrics. We report paired t-tests in \cref{fig:boxplot} and show that our method provides statistically significant improvements on all but the IHC TMA dataset. 

Overall, CoNSeP was the most challenging dataset for all the methods as shown by the lowest F1 score. This is likely due to the abundance of crowded, poorly resolved nuclei. InstanSeg performed especially well on this dataset, highlighting our methods robustness in challenging real world applications. Conversely, LyNSeC was the least challenging dataset, possibly due to the large number of annotations and low variability of images and nuclei shapes, where all methods performed highly. Furthermore, InstanSeg performed well on the small TNBC 2018 dataset, demonstrating its ability to learn from limited number of high quality annotations. On most datasets, our method's accuracy can be further improved using test-time augmentations (TTA), although the extra computational costs of applying TTA could be prohibitive in most real world applications and is not necessary for reaching state of the art accuracy.

\begin{figure}[h!]
    \centering
    \includegraphics[width=1\linewidth]{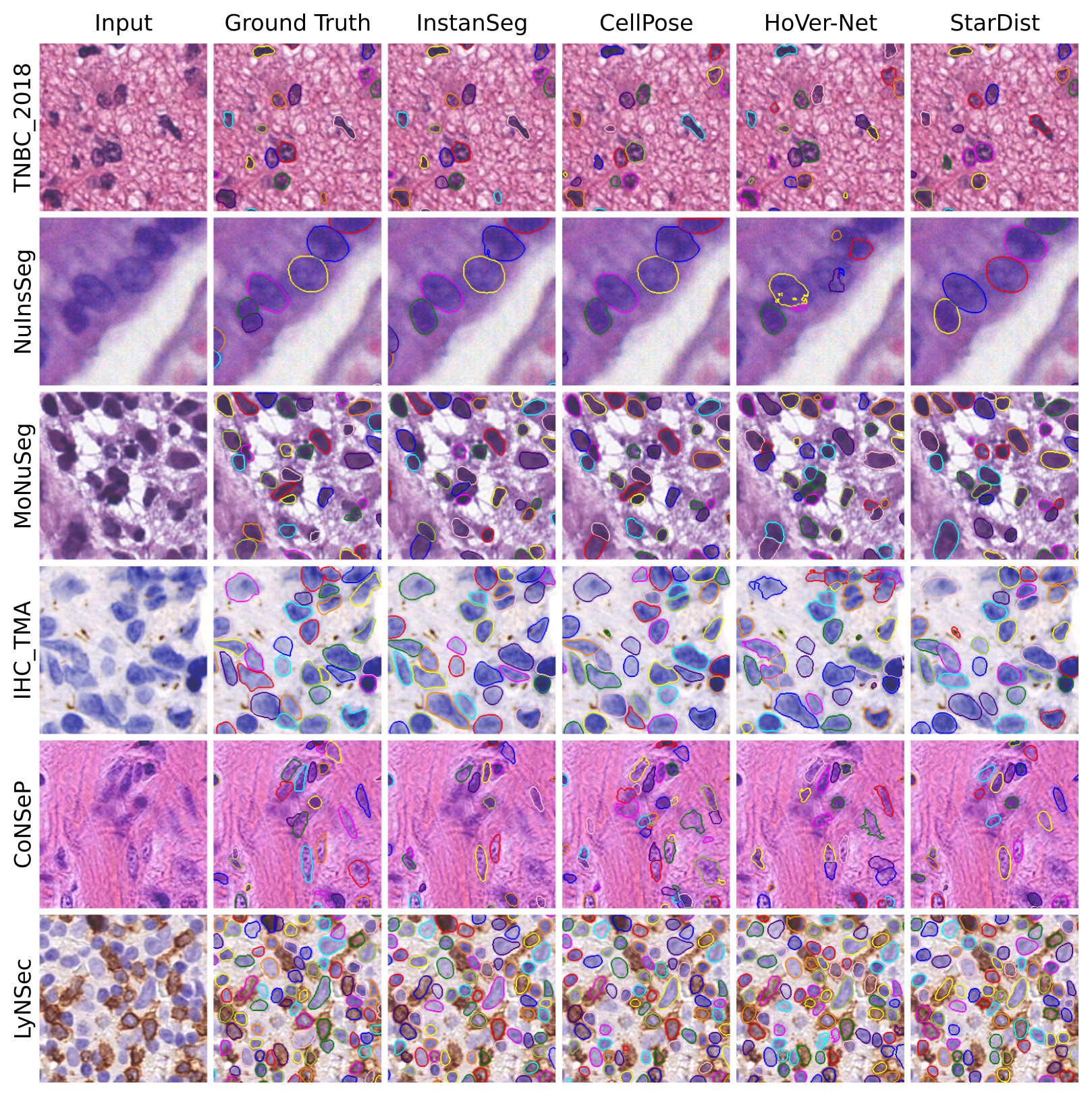}
    \caption{Qualitative segmentation results across six nucleus segmentation datasets. For each dataset, each of the four methods was retrained from scratch. The displayed images are $128 \times 128$ pixel crops which were randomly selected from the test set.}
    \label{fig:qualitative}
\end{figure}

We illustrate qualitative segmentation results in \cref{fig:qualitative} and show that detail in the object boundaries is mostly preserved in InstanSeg, while the instance boundaries of Stardist can only be approximated as star convex shapes. Hence, InstanSeg has the potential to provide additional cellular features of nuclear perimeter and some detail in the granularity of the nuclear envelope as compared to StarDist.

In our comparative analysis, Hover-Net exhibited lower performance across all datasets. This underperformance suggests the method may be less effective as a generalized out of the box solution for nucleus segmentation. Nevertheless, we acknowledge that the model may be improved by fine tuning training / testing hyperparameters on the individual datasets.

Some segmentation methods, including StarDist \citep{schmidt_cell_2018}, allow for the segmentation of overlapping objects, i.e, assigning single pixels to multiple objects. While, in theory, \textit{InstanSeg} could allow for the prediction of overlapping objects, we decide to flatten the output of the method by assigning pixels to the most probable of any overlapping instances. This decision was motivated by the fact that current annotated public datasets do not contain any overlapping objects, as well as the lack of computational tools for downstream analysis of overlapping cells in most current software.

Several public segmentation datasets do not report imaging pixel size (e.g. Data Science Bowl 2018 \citep{caicedo_nucleus_2019}), and a number of segmentation methods have treated image resolution as an unknown parameter that has to be estimated by the model \citep{stringer_cellpose_2021}. However, knowledge of accurate pixel size is not only crucial for accurate segmentation and most downstream analysis, it has implications on computational efficiency and memory requirements. Here, we assume that pixel size is recorded during the imaging process, and the image can be resized to a standard pixel size for use with an appropriate InstanSeg model. While this could limit InstanSeg's use for images where the pixel size is unknown, it provides a natural way to apply a model to similar images that have been acquired at different magnifications. It also reduces the risk of false detections arising from structures that may resemble nuclei in appearance but are a completely different size.

\subsection*{Ablations}

For investigating the effect of simplifying InstanSeg, we train on all six datasets rather than considering the datasets independently.  We report the results in \cref{tab:ablations}. We show that removing the conditional embeddings ($D_e = 0$) hinders segmentation accuracy, suggesting that the conditional embeddings can capture higher order information relating to instances. We also show that increasing the dimensionality of the positional embbeddings beyond the dimension of the image plane ($D_e > 2$) improves segmentation accuracy further.

\begin{table}[!h]
\centering
\caption{\label{tab:ablations} Effect of varying the depth of the positional embedding $D_e$ and the conditional embedding $D_p$. All six datasets were merged for this study.}
\begin{tabular}{|c|c|c|}
\hline
 & $F_1^{\mu}$ & $F_1^{0.5}$ \\ \hline
$D_e = 4$ $D_p = 4$ & 0.610 & 0.857 \\
$\hookrightarrow D_e = 4$ $D_p = 2$ & -0.006 & -0.006  \\
$\hookrightarrow D_e = 2$ $D_p = 2$ & -0.010 & -0.011  \\
$\hookrightarrow D_e = 0$ $D_p = 2$ & -0.020 & -0.014 \\
\hline
\end{tabular}
\end{table}

\subsection*{Time efficiency}

\begin{figure}
    \centering
    \includegraphics[width=1\linewidth]{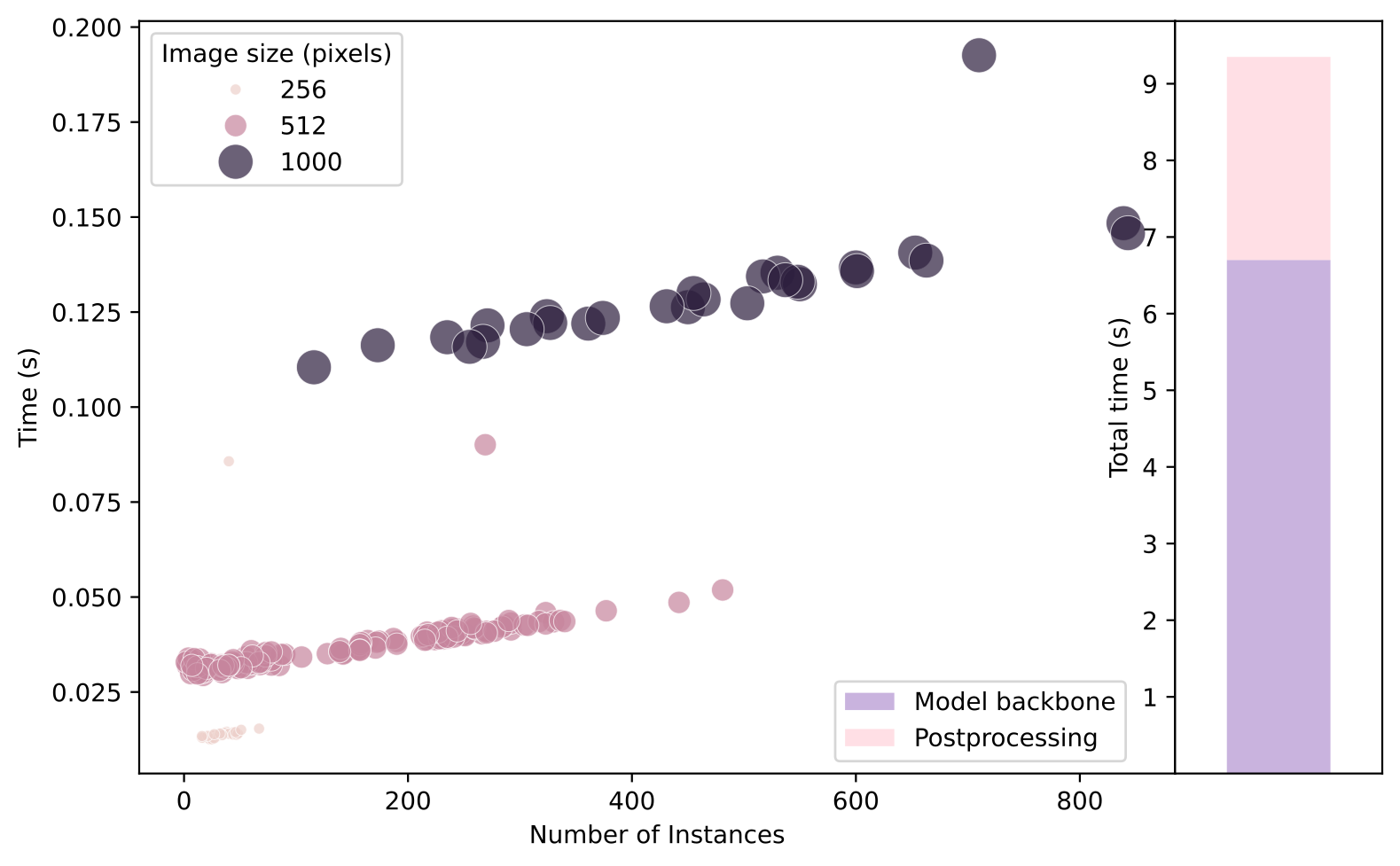}
    \caption{InstanSeg inference time is approximately linear with increasing number of detected instances and quadratic with the image dimensions. Timing performed on a laptop GPU. Unlike other methods, the conversion of model outputs to labeled instances is highly efficient, and applying the model backbone (UNet) is the efficiency bottleneck. Timing is for processing 199 images containing 36,073 instances.}
    \label{fig:timing}
\end{figure}

We profile InstanSeg on the combined test splits of all six datasets, totaling 199 images containing 36,073 instances. Profiling was performed on a laptop GPU (Quadro RTX 3000, 6GB), using mixed precision (FP16) and a fixed batch size of one. We report profiling results in \cref{fig:timing}. We compare InstanSeg's timing performance to four of the most widely implemented segmentation methods in \cref{tab:timing-efficiency}. We found that InstanSeg was nearly three times faster than the next fastest method StarDist and over ten times faster than CellPose. The high inference speed of InstanSeg is in large part owed to a highly parallelized GPU accelerated conversion of model outputs to labelled instances. Our postprocessing was over 23 times faster than EmbedSeg's iterative pixel clustering method. Furthermore, InstanSeg only required 1.5 GB of GPU memory, making it suitable for deployment on widely available hardware. 

\begin{table}[!h]
\centering
\caption{\label{tab:timing-efficiency}Time (in seconds) for processing 199 images containing 36,073 instances. Image sizes varied from $256 \times 256$ to $1000 \times 1000$ pixels. HoVer-Net model time is inflated as the method depends on tiling with large tile overlaps. All the methods were run on a laptop with a Quadro RTX 3000 GPU. StarDist is implemented in Tensorflow, whereas all the others are implemented in Pytorch.}
\small
\begin{tabular}{|l|c|c|c|}
\hline
& Model (s) & Postprocessing (s) & Total (s) \\
\hline
StarDist & 20.1 & 6.5 & 26.6 \\
HoVer-Net & 357.6 & 80.6 & 438.2 \\
CellPose & - & - & 103.2 \\
EmbedSeg & 9.4 & 63.3 & 72.7 \\
InstanSeg & \textbf{6.7} & \textbf{2.7} & \textbf{9.4} \\
\hline
\end{tabular}
\end{table}

\subsection{QuPath extension}

\begin{figure}
    \centering
    \includegraphics[width=1\linewidth]{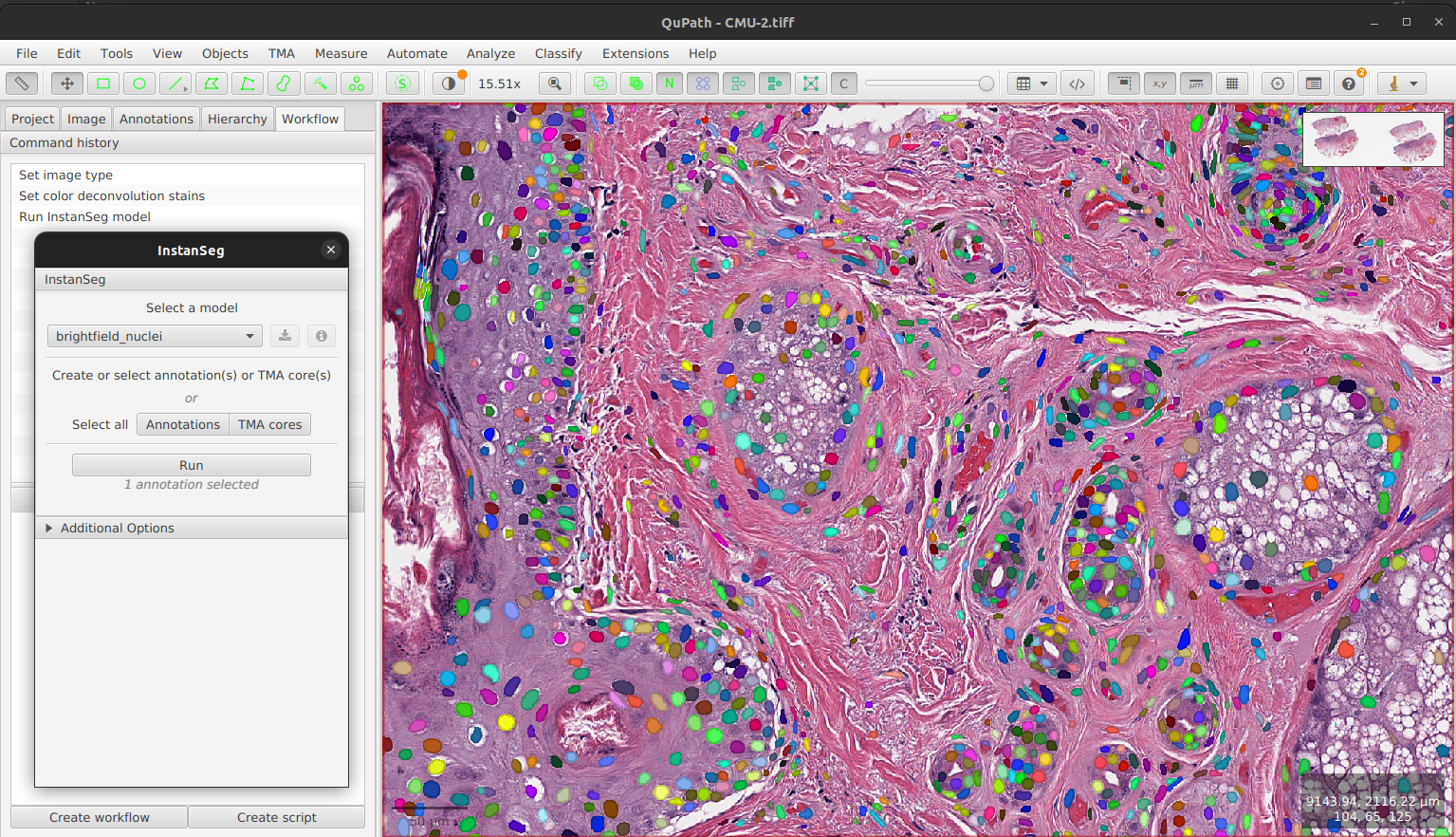}
    \caption{Screenshot showing the results of \textit{InstanSeg} extension in the QuPath software. QuPath enables InstanSeg to be integrated into full analysis pipelines with no coding experience required. The extension supports GPU acceleration on both NVIDIA and Apple hardware.}
    \label{fig:qupath}
\end{figure}

Unlike most other segmentation algorithms, InstanSeg can be compiled end-to-end using TorchScript, enabling the method to be run outside of the Python environment. We built a QuPath extension, providing a user friendly interface for running InstanSeg. This can greatly enhance the accessibility of the method to biologists, and enables \textit{InstanSeg} to be easily integrated in full analysis pipelines with limited coding experience. Our extension supports GPU acceleration on both NVIDIA and Apple hardware (\cref{fig:qupath}), and is available at this \href{https://github.com/instanseg/qupath-extension-instanseg}{https URL} \footnote{https://github.com/instanseg/qupath-extension-instanseg}.

\section{Conclusion}

We have proposed a novel method for the segmentation of cells in microscopy images and introduced the use of a neural network to cluster embeddings around optimally selected seed pixels. Our methodology provides several major improvements over previous state of the art embedding-based instance segmentation methods, including higher accuracy, speed and portability. Higher segmentation accuracies will allow for better quantification of cell properties and other downstream analyses, such as classifying cells and interrogating their spatial arrangement in tissue. Furthermore, the improved efficiency of InstanSeg enables the study of high-throughput imaging data from large image collections on widely available hardware. 

InstanSeg's improved portability enables its integration into end-to-end analysis pipelines through user-friendly and open-source software. Unlike most deep learning-based algorithms in the field, InstanSeg is not restricted to running through Python with GPU acceleration using CUDA; rather, it can also be used from other programming languages and includes GPU support on Apple Silicon. By demonstrating the benefits of packaging model inference and postprocessing steps in TorchScript, we hope that our example will encourage more computer vision researchers to develop portable methods that can be readily integrated into other software. Such efforts are crucial to enable standardization and wider adoption by domain experts.

While InstanSeg was designed and optimized for the segmentation of cells and nuclei, its approach can be applied to other instance segmentation tasks, both within and outside the biomedical domain. Relaxed assumptions on the shape of instances may make InstanSeg a good choice for segmenting complex, non-convex structures, while its high efficiency offer benefits for timelapse and video data. We are currently extending InstanSeg to distinguish cell compartments in highly-multiplexed images, as well as to segment different biological entities across a wider range of imaging modalities.

\textbf{Acknowledgements:}
TG and BP were supported by the United Kingdom Research and Innovation (grant EP/S02431X/1), UKRI Centre for Doctoral Training in Biomedical AI at the University of Edinburgh, School of Informatics. HB was supported by the EPSRC Visual AI grant EP/T028572/1. The development of the QuPath extension has been made possible in part by grant number 2021-237595 from the Chan Zuckerberg Initiative DAF, an advised fund of Silicon Valley Community Foundation. This research was funded in part by the Wellcome Trust 223750/Z/21/Z. For the purpose of open access, the author has applied a Creative Commons Attribution (CC BY) licence to any Author Accepted Manuscript version arising from this submission.
We extend our gratitude to the curators of the publicly available segmentation datasets we have referenced in this work.

\bibliographystyle{plainnat}

\bibliography{references.bib}
\end{document}